\documentclass[
]{ceurart}

\usepackage{caption,booktabs}
\usepackage{multirow}
\usepackage{siunitx}

\begin{document}

\copyrightyear{2021}
\copyrightclause{Copyright for this paper by its authors.
  Use permitted under Creative Commons License Attribution 4.0
  International (CC BY 4.0).}

\conference{Woodstock'21: Symposium on the irreproducible science,
  June 07--11, 2021, Woodstock, NY}

\title{One to rule them all: Towards Joint Indic Language Hate Speech Detection.}

\author[1]{Mehar Bhatia*}
\author[1]{Tenzin Singhay Bhotia*}
\author[1]{Akshat Agarwal}
\author[1]{Prakash Ramesh}
\author[1]{Shubham Gupta}
\author[1]{Kumar Shridhar}
\author[1]{Felix Laumann}
\author[1]{Ayushman Dash}
\address[1]{NeuralSpace, London, United Kingdom}

\begin{abstract}
  This paper is a contribution to the \textit{Hate Speech and Offensive Content Identification in Indo-European Languages} (HASOC) 2021 shared task. Social media today is a hotbed of toxic and hateful conversations, in various languages. Recent news reports have shown that current models struggle to automatically identify hate posted in minority languages. Therefore, efficiently curbing hate speech is a critical challenge and problem of interest. We present a multilingual architecture using state-of-the-art transformer language models to jointly learn hate and offensive speech detection across three languages namely, English, Hindi, and Marathi. On the provided testing corpora, we achieve Macro
  F1 scores of 0.7996, 0.7748, 0.8651 for sub-task 1A and 0.6268, 0.5603 during the fine-grained classification of sub-task 1B. These results show the  efficacy of exploiting a multilingual training scheme. 
\end{abstract}

\maketitle
\section{Introduction}

Since the proliferation of social media users worldwide, platforms like Facebook, Twitter, or Instagram have suffered from a rise of hate speech by individuals and groups. In a large-scale study on Twitter, and Whisper, \cite{silva2016analyzing} empirically show the prevalence of abusive comments and toxic language on such platforms and its most prevalent categories: race, physical and gender.

A Bloomberg article~\cite{bworld} reported that users have even found new ways of bullying others online using euphemistic emojis. Widespread use of such abusive language on social media platforms often causes public embarrassment to victims leading to major repercussions. Recently, Twitch filed a lawsuit against two users who targeted LGBTQ+ and Black streamers with hate speech~\cite{twitch}. One week later, content creators boycotted the game-streaming platform due to the inability to control the hateful content. Observing the growing usage of online hate, often anonymous, that comes at a scale unmanageable by human moderators, it is essential for social media platforms to control the abuse of users’ freedom of expression and maintaining an inclusive and respectful society. To enforce such supervision, platforms must be able to develop a system that can identify hate speech amongst billions of text comments by users at scale. 

\begin{figure}[ht]
  \centering
  \includegraphics[width=\linewidth]{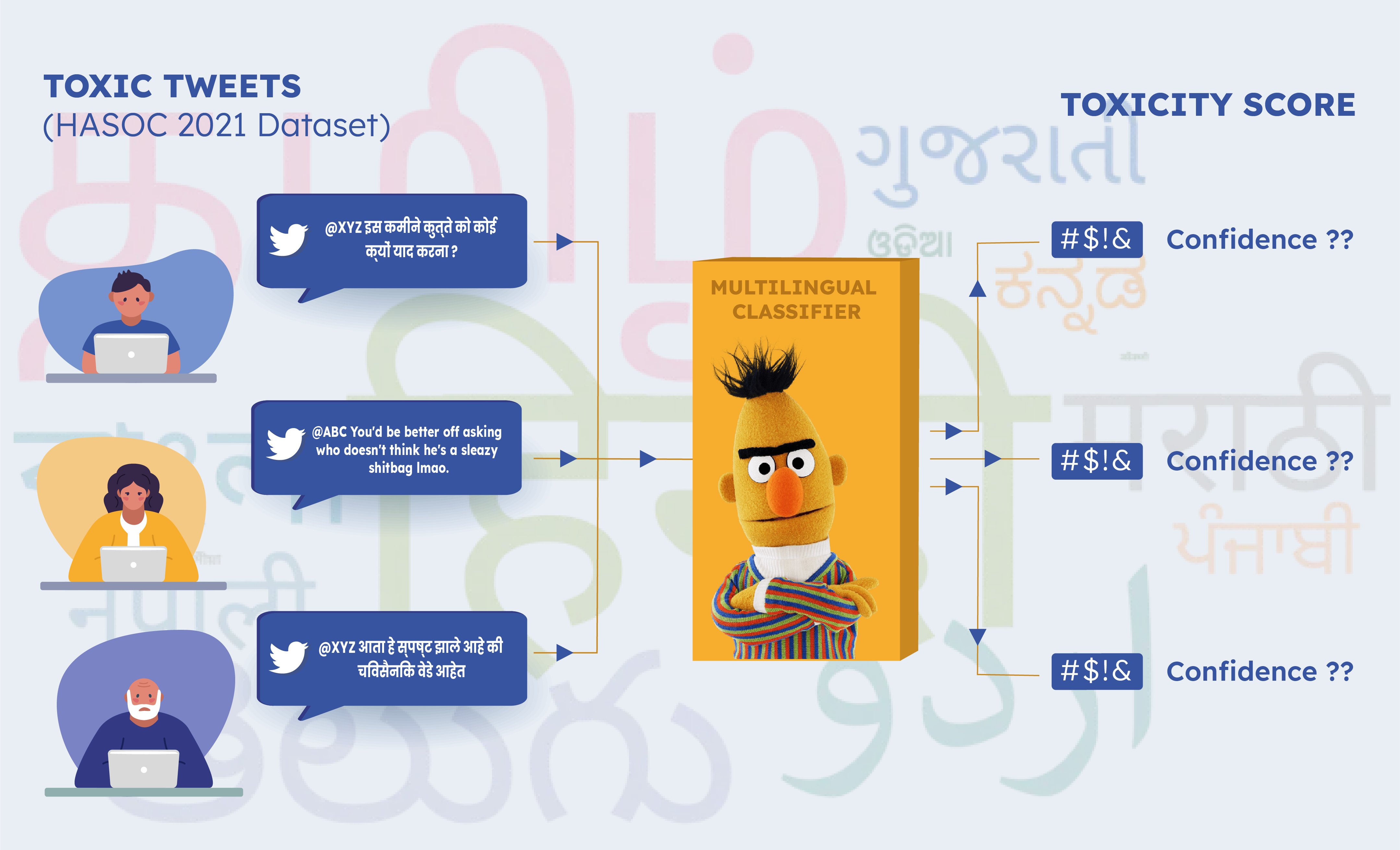}
  \caption{Overview of HASOC 2021 Problem Statement}
  \label{fig:problem}
\end{figure}

There have been research contributions in solving the problem of identifying abusive comments or other forms of toxic language~\cite{waseem2016hateful, watanabe2018hate, al2019detection, zhang2019hate}, however, most of them have majorly focused on high-resource languages, predominantly English. As social media connects people from all over the world, communicating in different languages, much of the potentially hateful content is present in a multilingual setting. The failure to pay attention to non-English languages is allowing such offensive speech to flourish. The lack of datasets and models for various low-resource languages has made the task of hate speech identification extremely difficult. In this paper, we present our findings on a subset of Indic low-resource languages.

The HASOC (Hate Speech and Offensive Content) 2021 challenge has been organized as a step towards this direction in three languages - English, Hindi, and Marathi. Figure~\ref{fig:problem} demonstrates HASOC 2021 problem statement. We focus on sub-task 1A and 1B of this competition, which we describe in the following paragraph.

Sub-task 1A focuses on hate speech and offensive language identification in English Hindi, and Marathi. It is a simple binary classification task in which participating systems are required to classify tweets into one of the two classes, namely: 
\begin{itemize}
    \item \textbf{(HOF)} \textit{Hate and Offensive:} Posts of this category contain either hate, offense, profanity, or a combination of them.
    \item \textbf{(NOT) }\textit{Non-Hate and offensive:} Posts of this category do not contain any hate speech, profane or offensive content.
\end{itemize}

Sub-task 1B is a multi-class classification task in English and Hindi. In this task, hate speech and offensive posts from sub-task A are further classified into the following three categories.
\begin{itemize}
    \item \textbf{(HATE)} \textit{Hate speech:} Posts of this category contain hate speech content.
    \item \textbf{(OFFN)} \textit{Offensive:} Posts of this category contain offensive content.
    \item \textbf{(PRFN)} \textit{Profane:} Posts of this category contain profane words.
\end{itemize}

In this paper, we make the following contributions: 
\begin{itemize}
    \item A pre-processing pipeline for modeling hate speech in the text of tweet domain.
    \item A joint fine-tuning procedure that empirically proves to outperform other approaches in hate speech detection.
    \item A summary of different approaches that did not work as expected.
    \item The implementation and idea behind our winning approach for one of HASOC 2021 sub-tasks.
\end{itemize}

In the forthcoming sections, we give a brief overview of past approaches as related work in section \ref{sec:rw}. Then, we present a detailed description of the statistics of datasets used in section~\ref{sec:dataset}. We present our approach in section~\ref{sec:approach}, delineating upon our pre-processing steps and model architecture. We highlight our model hyperparameters and other experimental details in section~\ref{sec:experiment}. Later, in section~\ref{sec:results}, we display our final results and elaborate on various other approaches that did not work well in section~\ref{sec:takeaways}. We end with our conclusion and point to future work in section~\ref{sec:conclusion}.

\section{Related Work}\label{sec:rw}

In the past, there have been many approaches to tackle the problem of hate speech identification. \citet{kwok2013locate} have experimented with a simple bag of words (BOW) approach to identify hate speech. While being light-weight, these models performed poorly with high false positive rates. Including various core natural language processing (NLP) features like part of speech tags~\cite{chen2012detecting} and N-gram graphs~\cite{themeli2018hate} have helped in improving the performance.
Lexical methods using TF-IDF and SVM as a classification model have achieved surprisingly good performance~\cite{reddy2020dlrg}.

With the rise of embedding words in distributed representations, researchers have leveraged word embeddings like Glove~\cite{pennington2014glove}, and FastText~\cite{bojanowski2017enriching} for embedding discrete text into a latent space and have improved the performance over standard BOW and lexical approaches. 

Recurrent Neural Networks (RNNs) for many years were the de-facto approach for tackling any natural language problem. The winning approach at the 2020 HASOC competition for Hindi~\cite{raja2021nsit} used a one-layer BiLSTM with FastText embeddings to identify hate speech. Similarly for English, the most accurate model~\cite{mishraa2020iiit_dwd} used an LSTM with Glove embeddings to represent text inputs. \citet{mohtaj2020tub} also used a character-based LSTM following a similar trend.

In recent times, self-attention-based transformer~\cite{vaswani2017attention} models and language models derived from its huge corpus trained encoders like BERT~\cite{devlin2018bert} have shown more promise than standard RNNs for most of the NLP tasks.
Many researchers have found BERT-like models to perform much better than other approaches majorly due to their high transfer learning prowess~\cite{mozafari2019bert}.
While there has been a lot of research on hate speech in general, experiments especially focusing on low-resource languages are less popular. Simple logistic regression using LASER embeddings has been shown to perform better than BERT-based models~\cite{aluru2020deep} indicating the need for more accurate multilingual base language models. Since then, we have witnessed the rise of multilingual language models like XLM-Roberta~\cite{conneau2019unsupervised}. In the following sections, we will delineate our approach of building a solution using XLM-Roberta for identifying hate speech along with an exhaustive comparison to other approaches.

\section{Dataset Description} \label{sec:dataset}
Datasets for HASOC 2021 \cite{hasoc2021mergeoverview} for English \cite{hasoc2021overview}, Hindi \cite{hasoc2021overview}, and Marathi \cite{gaikwad2021cross} languages were collected from social media platforms and comprises two sub-tasks. We focus only on the first sub-task which is divided into two, i.e., sub-task 1A and sub-task 1B. As shown in Table~\ref{tab:desciption}, each dataset instance consists of a unique \textit{hasoc\_id}, a \textit{tweet\_id}, full \textit{text} of the tweet, and target variables \textit{task\_1} and \textit{task\_2} for the sub-task 1A and 1B respectively. sub-task 1A is a binary classification problem with two target classes namely, \textbf{HOF} \textit{(Hate and Offensive)} and \textbf{NOT} \textit{(Non-Hate-offensive)}, whereas sub-task 1B is a further fine-grained classification. The data is further classified into four classes, namely \textbf{OFFN} \textit{(Offensive)}, \textbf{PRFN} \textit{(Profane)}, \textbf{HATE}, and \textbf{NONE} class. sub-task 1A challenges with datasets in English, Hindi, and Marathi languages, whereas only English and Hindi datasets are available for sub-task 1B.
The statistics of both the train and test data and shown in Table \ref{tab:train} and Table \ref{tab:test}. 

It can be seen that the datasets are highly imbalanced. For sub-task 1A, we notice that the number of hate and offensive tweets is almost double for English and Marathi. On the other hand, the number of non-hate-offensive tweets is 55\% higher for the Hindi dataset. In addition to that, for sub-task 1B, offered in English and Marathi language, the further classification of hate and offensive tweets are imbalanced in opposite ratios. 
\begin{table}[ht]
  \caption{Dataset Description}
  \label{tab:desciption}
  \begin{tabular}{cc}
    \toprule
    Columns & Description\\
    \midrule
    hasoc\_id & unique hasoc ID for each tweet\\
    tweet\_id & unique value for each tweet\\
    text & full text of the tweets\\
    task\_1 & target value for sub-task 1A (HOF or NOT)\\
    task\_2 & target value for sub-task 1B (OFFN, PRFN, HATE or NONE)\\
    
  \bottomrule
\end{tabular}
\end{table}

\begin{table}[ht]
  \caption{Class division of both sub-tasks for Train Dataset}
  \label{tab:train}
  \begin{tabular}{l|cc|cccc|c}
    \toprule
    {Language} & \multicolumn{2}{c|}{\textbf{Task 1}} & \multicolumn{4}{c|}{\textbf{Task 2}} & {\textbf{TOTAL}} \\
    {} & {HOF} & {NOT} & {HATE} & {OFFN} & {PRFN} & {NONE} & {} \\
    \midrule
    \textbf{English} & 2501 & 1342& 683 & 622 & 1196 & 1342 & 3843\\
    \textbf{Hindi} & 1433 & 3161 & 566 & 654 & 213 & 3161 & 4594\\
    \textbf{Marathi} & 1205 & 669 & $-$ & $-$ & $-$ & $-$ & 1874 \\
    \bottomrule
  \end{tabular}
\end{table}

\begin{table}[ht]
  \caption{Class division of both sub-tasks for Test Dataset}
  \label{tab:test}
  \begin{tabular}{l|cc|cccc|c}
    \toprule
    {Language} & \multicolumn{2}{c|}{\textbf{Task 1}} & \multicolumn{4}{c|}{\textbf{Task 2}} & {\textbf{TOTAL}} \\
    {} & {HOF} & {NOT} & {HATE} & {OFFN} & {PRFN} & {NONE} & {} \\
    \midrule
    \textbf{English} & 798 & 483 & 224 & 195 & 379 & 483 & 1281\\
    \textbf{Hindi} & 505 & 1027 & 215 & 215 & 44 & 1027 & 1532\\
    \textbf{Marathi} & 483 & 418 & $-$ & $-$ & $-$ & $-$ & 901\\
    \bottomrule
  \end{tabular}
\end{table}

\section{Approach} \label{sec:approach}
In this section, we demonstrate our approach of solving HASOC 2021 sub-task1A and sub-task1B tasks.

\subsection{Preprocessing}
For preprocessing the tweet data and hashtags, we used python libraries, namely tweet-preprocessor\footnote{https://github.com/s/preprocessor} and ekphrasis\footnote{https://github.com/cbaziotis/ekphrasis}, a segmenter built on Twitter corpus. For English data, the tweet-preprocessor's clean functionality was used to extract clean, parse and tokenize the tweet texts. For Hindi and Marathi data, we first tokenized the tweet text on whitespaces and symbols including colons, commas, semicolons, dashes, and underscores. Secondly, we used the tweet-preprocessor python library for the removal of URLs, hashtags, mentions, emojis, smileys, numbers, and reserved words (such as @RT which stands for Retweets). We also noticed the usage of words in English and Arabic in the Hindi and Marathi datasets. We first transliterated this text to the desired language using NeuralSpace's transliteration tool \footnote{https://docs.neuralspace.ai/transliteration/overview}. Later, if English or Arabic occurrences remain, we used python library langdetect \footnote{https://pypi.org/project/langdetect/} (a re-implementation of Google’s language-detection library \footnote{https://github.com/shuyo/language-detection} from Java to Python) to extract the pure Hindi and Marathi text within the tweet. 

\subsection{Feature Extraction} 
\label{section: feature_extraction}
To extract features for our classifier, we use tweet-preprocessor to supply various information fields, in addition to the cleaned content. The first feature is obtained from the hashtag text which is segmented into constituent and meaningful tokens using the ekphrasis segmenter. \textit{Ekphrasis} tokenizes the text based on a list of regular expressions. For example, the hashtags \textit{`\#\#JitegaModiJitegaBharat'}, \textit{`\#IPL2019Final'}, \textit{`\#hogicongresskijeet'} is tokenized to \textit{`Jitega Modi Jitega Bharat'}, \textit{`IPL 2019 Final'}, \textit{`hogi congress ki jeet`}.  Other features are  acquired from URLs within the text, name mentions such as \textit{`\@BJP4Punjab'}, \textit{`\@aajtak'}, \textit{`\@PMOIndia'}, and \textit{`\@narendramodi'}, and smileys and emojis. The extracted emojis were processed in two ways. 

First, we use emot\footnote{https://github.com/NeelShah18/emot} python library to obtain the textual description of a particular emoji in the text. Emot uses advanced dynamic pattern generation. For example, `rofl emoji' refers to `rolling-on-the-floor-and-laughing face' and `speak-no-evil emoji' refers to `speak-no-evil Monkey'. However, we felt this mapping is not sufficient as it does not highlight the genuine meaning of what the emoji represents in reality. Given that the usage of such emojis is so prevalent and that most of them inherently have emotions built-in, emojis can give a lot of insights into the sentiment of online text. For this reason, we also considered emoji2vec \cite{eisner2016emoji2vec} embeddings for 1661 emoji Unicode symbols learned from a total of 6088 descriptions in the Unicode emoji standard. Previous work has demonstrated the usefulness of this by evaluating various tasks such as Twitter sentiment analysis \cite{eisner2016emoji2vec}. For example, consider `pray emoji' and `tipping-hand-woman emoji', which map to `the-folded-hands' symbol and the `woman-tipping-hand' emoji. The textual representation will not showcase the emoji’s association with `showing gratitude, expressing an apology, sentiments such as hope or respect or even a high five` which is its real-world implication. On the other hand, the person-tipping symbol is commonly used to express \textit{`sassiness'} or sarcasm. We expect emoji2vec to capture these kinds of analogy examples.

\subsection{Proposed Architecture}
We leverage Transformer-based \cite{vaswani2017attention} masked language models to generate semantic embeddings for the cleaned tweet text. 

We use the available training corpora and fine-tune the transformer layers in a multilingual fashion for our downstream task. We experimented with various multi-lingual transformer models, i.e XLM-RoBERTa (XLMR), mBERT(multilingual BERT), and DistilmBERT (multilingual-distilBERT). A summary for each model is as follows:
\begin{itemize}
    \item \textbf{XLM-RoBERTa: }The pre-training of XLM-RoBERTa is based on 100 languages, using around 2.5TB of preprocessed CommonCrawl dataset to train cross-language representations in a self-supervised manner. XLM-RoBERTa \cite{conneau2019unsupervised} shows that the use of large-scale multi-language pre-training models can significantly improve the performance of cross-language migration tasks.
    \item \textbf{mBERT: }Multilingual BERT \cite{devlin2018bert} uses Wikipedia data of 102 languages, totaling to 177M parameters, and is trained using two objectives i.e, 1) using a masked language modeling (MLM) when 15\% of input is randomly masked, and 2) using next sentence prediction.
    \item \textbf{DistilmBERT: }Distil multilingual BERT \cite{sanh2019distilbert}is a distilled version of the above mBERT model. It is also trained on the concatenation of Wikipedia in 102 different languages. It has a total of 134M parameters. On average DistilmBERT is twice as fast as mBERT-base.
\end{itemize}

To solve the sub-task 1A of three languages (English, Hindi, and Marathi), and sub-task 1B of two languages (English and Hindi) at the same time, we adopt these multi-lingual models.

As mentioned in Section~\ref{section: feature_extraction}, we generate semantic vector representations for all the emojis and smileys, their respective text, and segmented hashtags within the tweet. We encode the emoji, smiley text embeddings, and hashtag embeddings in the same latent space. To create the emojis' semantic embeddings, emoji2vec is utilized. An important point to notice is that the segmented hashtags and text descriptions of emojis can be of variable length. Hence, we generate the centralized emoji or hashtag
representation by averaging the vector representations. This is a simple approach proposed by \cite{arora2016simple} to produce a comprehensive vector representation for sentences. 

\begin{figure}
  \centering
  \includegraphics[width=\linewidth]{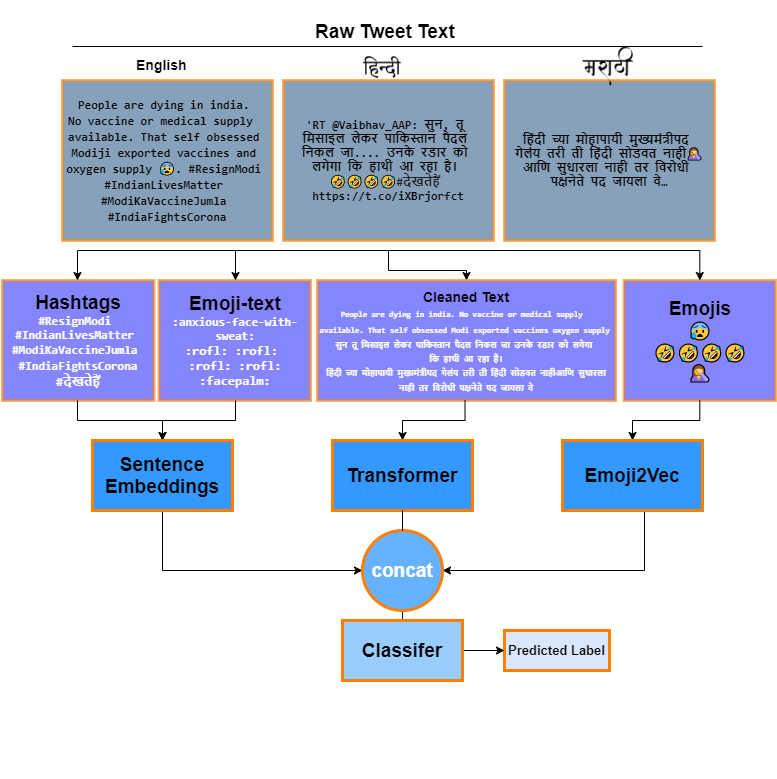}
  \caption{High level overview of our proposed architecture}
\end{figure}

\section{Experimental Details} \label{sec:experiment}
We used Hugging Face’s implementation and corresponding pre-trained models of XLM-RoBERTa \footnote{}, multilingual BERT\footnote{https://huggingface.co/bert-base-multilingual-uncased}, and multilingual-distilBERT \footnote{https://huggingface.co/distilbert-base-multilingual-cased} in our proposed architecture. Our architectures using Transformer models with custom classification heads were implemented using PyTorch. We used Adam optimizer for training with an initial learning rate of 2e-4, dropout probability of 0.2 with other hyper-parameters set to their default values. We use a cross-entropy loss to update the weights. We have also used UKPLab’s sentence-transformers library \footnote{https://github.com/UKPLab/sentence-transformers} to encode the hashtags and textual descriptions of the emojis.

All the fine-tuned language models broadly fall into two following categories.
\begin{itemize}
    \item Monolingual: These are a type of model that have been fine-tuned on only the respective target language. For instance, we only use the English train dataset to fine-tune the model and then infer on the English test set only.
    \item Multilingual: These are a type of model that has been fine-tuned on a combination of all available languages irrespective of the target language. For instance, to train a model for the English target language on sub-task 1A, we combine the train datasets for all languages (English, Hindi, and Marathi) and then fine-tune the model once. Such a model can then be inferenced on any given target language. Intuitively, such a training scheme provides three benefits.    \begin{itemize}
        \item It enforces joint modeling of the training distribution for all the given languages. Empirically we find this to perform better than individually modeling on respective language. 
        \item During inference, we only rely on one model to infer instead of a unique model for each language. An approach that can be extremely computed efficient for production.
        \item We combine naturally occurring human-annotated data just in a different language. Hence, it becomes a promising approach towards resolving poor model performance due to the data scarcity issue for low-resource languages.
    \end{itemize}
\end{itemize}
As shown in Table \ref{tab:results}, we empirically observe that a multilingual training setting clearly outperforms the monolingual setting across both the tasks in all three languages irrespective of the base model. For English sub-task 1, only mBERT and DistilmBERT score below the monolingual setting, but the difference is not as significant. This experiment suggests that multilingual training can be an important approach to train better-performing models, especially as it provides a step towards resolving the data scarcity issue for low-resource languages. It will be interesting to validate the generalizability of this hypothesis on different NLP tasks in the future.

All experiments were carried on a workstation with one NVIDIA A100-SXM4-40GB GPU with 12 CPU cores. We used a batch size of 64 throughout. For the initial experiments, we have divided the
released training data into a training set and  a validation set and conducted the experiments using accuracy as the performance metric. Finally, the performance of the proposed system was tested on the test set released by the organizers. For these experiments, we combined all the training and validation data into a single training set and applied our algorithm. For the multilingual setting, our experiment took 3.5 hours to train till convergence. For the monolingual setup, our model took 1.2 hours to train till convergence. 
\section{Results} \label{sec:results}
Table \ref{tab:results} presents our performance evaluations of our proposed architectures on the provided testing set. The evaluation metric used throughout is the macro F1-score. 

We see that fine-tuned multilingual models clearly beat the baselines by at least \% , demonstrating the prowess of large language models. 

It is observed from Table \ref{tab:results}, that for all three languages, XLM-RoBERTa has outperformed similar multilingual Transformer models such as mBERT (multilingual BERT) and distilmBERT (multilingual-distilBERT) on our hate speech detection task. We observe a minimum absolute gain of 1.63 F1 and 1.20 F1 for Task1 and Task 2 respectively via the multilingual approach with XLM-RoBERTa. While a maximum absolute gain of 2.1 F1 and 2.31 F1 for Task1 and Task 2 respectively. Empirically such significant improvements suggest the importance of multilingual training over monolingual training.
Notably, multilingually trained XLM-RoBERTa secured the 1st position among 24 participants and the 5th position among 34 participants on the HASOC 2021 leaderboard for Task 1 and 2 respectively. Securing such high ranks again indicates the importance of the multilingual approach and calls for a detailed investigation of this approach on other tasks as well for future work.

\begin{table*}
  \caption{Results of proposed architectures on test data}
  \label{tab:results}
  \begin{tabular}{ccccccl}
    \toprule
    \multicolumn{2}{c}{Proposed Architecture} & \multicolumn{2}{c}{\textbf{English}} & \multicolumn{2}{c}{\textbf{Hindi}} & \multicolumn{1}{c}{\textbf{Marathi}}\\
    \cmidrule{1-2} \cmidrule{3-4} \cmidrule{5-7} 
    \textbf{Model} & \textbf{Mono/Multi}& \textbf{Task1} & \textbf{Task2} & \textbf{Task1} & \textbf{Task2} & \textbf{Task1}\\
    \midrule
    XLM-R Finetuned & Monolingual & 0.7786 & 0.6148 & 0.7585 & 0.5447 & 0.8420\\
    mBERT Finetuned &Monolingual & 0.7631 & 0.5703 &0.7462&0.5202&0.8269\\
    DistilmBERT Finetuned &Monolingual & 0.7637 & 0.5826 &0.7554&0.5267&0.8350\\
    \hline
    \textbf{XLM-R Finetuned} & \textbf{Multilingual} & \textbf{0.7996} & \textbf{0.6268} & \textbf{0.7748} & \textbf{0.5603} & \textbf{0.8651}\\
    mBERT Finetuned &Multilingual & 0.7626 & 0.6121 &0.7593&0.5554&0.8461\\
    DistilmBERT Finetuned &Multilingual & 0.7592 & 0.6104 &0.7589&0.5591&0.8510\\
  \bottomrule
\end{tabular}
\end{table*}

\section{Key Takeaways} \label{sec:takeaways}
In this section, we aim to provide a checklist of various approaches and techniques that we implemented for this task but did not result in competitive positions on the leaderboard. We believe that our readers will benefit from this checklist during future work. 

To begin with, as the dataset was overall highly imbalanced across all languages, we perform SOUP (Similarity-based Oversampling and Undersampling processing), a technique in which the number of the minority class samples is increased the number of majority class samples are decreased to obtain a balanced data set. This technique was suggested by \cite{reddy2020dlrg} and we use this balanced data for performing the classification task, however, when compared to our best performing model, we see a drop of 5\% in accuracy.

Secondly, to add more training samples to our multilingual dataset, we use data augmentation techniques such as back-translation to generate this synthetic data. We adopt ML Translator API, which is Google’s Neural Machine Translation (NMT) system. This translation method was widely used because of its simplicity and zero-shot translation. With this method, we increase our dataset size by three times, however, we don't see any performance gains using this augmented dataset for our proposed architecture. Moreover, we observed a reduction of toxicity upon using this back-translation method possibly resulting in false labels for many instances. 

Based on winning approaches from \cite{mandl2020overview} and \cite{mandl2019overview}, we applied different machine learning algorithms, i.e, random forest, and LightGBM, a gradient boosting framework based on decision trees. These techniques have shown an average drop of 5.3\%. We also looked into two different deep neural networks approaches and tested them for all three languages. For the English model, we used GloVe \footnote{https://nlp.stanford.edu/projects/glove/} embeddings \cite{pennington2014glove} for both sub-tasks. This embedding layer is fed to a CNN model. The architecture comprises two convolutional, two dropouts, and two max-pooling layers accompanied by a flatten layer and a dense layer. We achieved a macro F1 score of 0.75 and 0.56 respectively on HASOC 2021 sub-task A and B test sets. For Hindi and Marathi models, we used fastText \footnote{https://fasttext.cc/docs/en/crawl-vectors.html} embeddings \cite{bojanowski2017enriching} for both the sub-tasks. Here, the embeddings are passed through a bi-directional LSTM model and a dropout layer, followed by a dense layer. We achieved macro F1 scores of 0.74, 0.54, and 0.84 on Hindi Task1, 2 and Marathi Task 1, respectively. Overall, we see that our final proposed architecture performed the best overall tasks.   

\section{Conclusion} \label{sec:conclusion}
This work has been submitted to CEUR 2021 Workshop Proceedings for the task, Identification of Hate and Offensive Speech in
Indo-European Languages (HASOC 2021). In this research, the problem of identifying hate and offensive content in tweets has been experimentally studied on three different language datasets namely, English, Hindi, and Marathi. We propose a joint language training approach based on recent advances in large-scale transformer-based language models and demonstrate our best results. We plan to further explore other novel methods of capturing social media text semantics as part of future work. We also aim to look at more accurate data augmentation techniques to handle the data imbalance and enhancing hate and offensive speech detection in social media posts.

\bibliography{main}

\appendix

\end{document}